\documentclass[10pt,twocolumn,letterpaper]{article}

\usepackage{wacv}
\usepackage{times}
\usepackage{epsfig}
\usepackage{graphicx}
\usepackage{amsmath}
\usepackage{amssymb}

\usepackage{color}
\usepackage{colortbl}
\usepackage{boldline}
\usepackage{hhline}
\usepackage{arydshln}
\usepackage{mathtools}
\usepackage{graphicx}
\usepackage{color}
\usepackage{ifthen}
\usepackage{multirow}
\usepackage{arydshln}
\makeatletter
\@namedef{ver@everyshi.sty}{}
\makeatother
\usepackage{tikz}
\usepackage{subfigure}

\usepackage[pagebackref=true,breaklinks=true,letterpaper=true,colorlinks,bookmarks=false]{hyperref}

\wacvfinalcopy 

\def\wacvPaperID{537} 

\ifwacvfinal\fi
\begin{document}

\title{SfMLearner++: Learning Monocular Depth \& Ego-Motion using Meaningful Geometric Constraints}

\author{Vignesh Prasad* \hspace{2cm} Brojeshwar Bhowmick\\
Embedded Systems and Robotics, TCS Innovation Labs, Kolkata, India\\
{\tt\small \{vignesh.prasad,b.bhowmick\}@tcs.com}
}

\maketitle
\ifwacvfinal\thispagestyle{empty}\fi

\begin{abstract}
   Most geometric approaches to monocular Visual Odometry (VO) provide robust pose estimates, but sparse or semi-dense depth estimates. Off late, deep methods have shown good performance in generating dense depths and VO from monocular images by optimizing the photometric consistency between images. Despite being intuitive, a naive photometric loss does not ensure proper pixel correspondences between two views, which is the key factor for accurate depth and relative pose estimations. It is a well known fact that simply minimizing such an error is prone to failures.\par
   
   We propose a method using Epipolar constraints to make the learning more geometrically sound. We use the Essential matrix, obtained using Nist\'er's Five Point Algorithm, for enforcing meaningful geometric constraints on the loss, rather than using it as labels for training. Our method, although simplistic but more geometrically meaningful, using lesser number of parameters, gives a comparable performance to state-of-the-art methods which use complex losses and large networks showing the effectiveness of using epipolar constraints. Such a geometrically constrained learning method performs successfully even in cases where simply minimizing the photometric error would fail.\par

\end{abstract}
\vspace{-1em}
\section{Introduction}
\label{sec:intro}

Off late, the problem of dense monocular depth estimation and/or visual odometry estimation using deep learning based approaches has gained momentum. These methods work in a way similar to the way in which we as humans develop an understanding based on observing various scenes which have consistencies in structure. Such methods, including the one proposed in this paper, are able to make structural inferences based on similarities in the world and make relative pose estimates based on this and vice versa. Some of them either use depth supervision \cite{eigen2014depth,hoiem2005automatic,duggal2016hierarchical,saxena2007learning} or pose supervision using stereo rigs \cite{garg2016unsupervised,godard2017unsupervised} or ground truth poses for relative pose estimation \cite{saxena2017exploring} or both pose and depth supervision among others \cite{ummenhofer2017demon}. In SE3-Nets\cite{byravan2017se3}, the authors estimate rigid body motions directly from pointcloud data.  \par

In order to deal with non-rigidity in the scene, SfMLearner\cite{zhou2017unsupervised}, predicts an "explainability mask" along with the pose and depth in order to discount regions that violate the static scene assumption. This can also be done by explicitly predicting object motions and incorporating optical flow as well, such as in SfM-Net\cite{vijayanarasimhan2017sfm} and GeoNet\cite{yin2018geonet}. Additional efforts to ensure consistent estimates are explored by Yang \textit{et al.}\cite{yang2017unsupervised} using depth-normal consistency. \par

Most of the above-mentioned works incorporate various complicated loss functions rather than explicitly leveraging 3D geometric constraints. In order to ensure better correspondences, thereby better geometric understanding, we propose a method that uses Epipolar constraints to help make the learning more geometrically sound. This way we constrain the point correspondences to lie on their corresponding epipolar lines. We make use of the Five Point algorithm to guide the training and improve predictions.  We use the Essential matrix, obtained using Nist\'er's Five Point Algorithm \cite{nister2004efficient}, for enforcing meaningful geometric constraints on the loss, rather than using it as labels for training. We do so by weighing the losses using epipolar constraints with the Essential Matrix. This helps us account for violations of the static scene assumption, thereby removing the need to predict explicit motion masks, and also to tackle the problem of improper correspondence generation that arises by minimizing just the photometric loss.\par 

The key contribution is that we use a geometrically constrained loss using epipolar constraints, which helps discount ambiguous pixels, thereby allowing us to use lesser no. of parameters. Our proposed method results in more accurate depth images as well as reduced errors in pose estimation. Such a geometrically constrained learning method performs successfully even in cases where simply minimizing the photometric error fails. We also use an edge-aware disparity smoothness to help get sharper outputs which are comparable to state-of-the-art methods that use computationally intesive losses and a much larger no. of parameters compared to us.\par 

\section{Background}
\label{sec:bg}

\subsection{Structure-from-Motion (SfM)}
\label{ssec:sfm}
\textbf{Structure-from-Motion} refers to the task of recovering 3D structure and camera motion from a sequence of images. It is an age-old problem in computer vision and various toolboxes that perform SfM have been developed \cite{wu2013towards,furukawa2010towards,sturm1996factorization}. 
Traditional approaches to the SfM problem, though efficient, require accurate point correspondences in computing the camera poses and recovering the structure. A spinoff of this problem comes under the domain of Visual SLAM or VO, which involves real-time estimation of camera poses and/or a structural 3D map of the environment. There approaches could be either sparse\cite{mur2015orb,klein2007parallel,forster2014svo,maity2017edge,jose2015realtime}, semi-dense\cite{engel2014lsd,engel2018direct} or dense\cite{newcombe2011dtam,alismail2016enhancing}. Both methods suffer from the same sets of problems, namely improper correspondences in texture-less areas, or if there are occlusions or repeating patterns. While approaching the problem in a monocular setting, sparse correspondences allow one to estimate depth for corresponding points. However estimating dense depth from a single monocular image is a much more complex problem.  \par 
\subsection{Epipolar Geometry}
\label{ssec:epipolar}
We know that a pixel $p$ in an image corresponds to a ray in 3D, which is given by its normalized coordinates $\Tilde{p} = K^{-1}p$, where $K$ is the intrinsic calibration matrix of the camera. From a second view, the image of the first camera center is called the epipole and that of a ray is called the epipolar line. Given the corresponding pixel in the second view, it should lie on the corresponding epipolar line. This constraint on the pixels can be expressed using the Essential Matrix $E$ for calibrated cameras. The Essential Matrix contains information about the relative poses between the views. Detailed information regarding normalized coordinates, epipolar geometry, Essential Matrix etc, can be found in \cite{hartley2003multiple}.\par
Given a pixel's normalized coordinates $\Tilde{p}$ and that of its corresponding pixel in a second view $\hat{\Tilde{p}}$, the relation between $\Tilde{p}$, $\hat{\Tilde{p}}$ and $E$ can be expressed as:

\vspace{-0.8em}
\begin{equation}
\label{eq:epipolar}
\hat{\Tilde{p}}^TE\Tilde{p} = 0
\vspace{-0.2em}
\end{equation}

Here, $E\Tilde{p}$ is the epipolar line in the second view corresponding to a pixel $p$ in the first view. In most cases, there could be errors in capturing the pixel $p$, finding the corresponding pixel $\hat{p}$ or in estimating the Essential Matrix $E$. Therefore in most real world applications, rather than imposing Eq. \ref{eq:epipolar}, the value of $\hat{\Tilde{p}}^TE\Tilde{p}$ is minimized in a RANSAC\cite{fischler1981random} like setting. We refer to this value as the epipolar loss in the rest of the paper. \par 

We refer to a pixel's homogeneous coordinates as $p$ and the normalized coordinates as $\Tilde{p}$ in the rest of the paper. We also refer to a corresponding pixel in a different view as $\hat{p}$ and in normalized coordinates as $\hat{\Tilde{p}}$.


\subsection{Depth Image Based Warping}
\label{ssec:dibr}
\vspace{-0.5em}
Novel view synthesis by warping the input image to a new pose is an important step in understanding the geometry of a scene. In order for the model to have a decent understanding of scene geometry, the transformed image needs to be consistent with the image from that location. This is the key idea behind minimizing the photometric error. One of the most common approached to doing this is by using differentiable warping and bi-linear sampling \cite{jaderberg2015spatial} which have been in use for a variety of applications like learning optical flow \cite{jason2016back}, video prediction \cite{patraucean2015spatio} and image captioning \cite{johnson2016densecap}. 
This concept has been applied in works along similar lines where bi-linear sampling is used to obtain a resulting image after warping a view using scene depth and the relative transformation between two views. \par
In this approach, given a warped pixel $\hat{p}$, its pixel value $I_s(\hat{p})$ is interpolated using 4 nearby pixel values of $\hat{p}$ (upper left and right, lower left and right) i.e. $\hat{I}_s(p) = I_s(\hat{p}) = \sum_{i} \sum_{j} w_{ij} I_s(p_{ij})$
where $i\in\{\lfloor\hat{p}_x\rfloor,\lceil\hat{p}_x\rceil\}$, $j\in\{\lfloor\hat{p}_y\rfloor,\lceil\hat{p}_y\rceil\}$ and $w_{ij}$ is directly proportional to the proximity between $\hat{p}$ and $p_{ij}$ such that $\sum w_{ij} = 1$. Further explanation regarding this can be found in \cite{jaderberg2015spatial}.\par

\section{Proposed Approach}
\label{sec:approach}

\begin{figure*}[t]
	\begin{center}
		\includegraphics[width=0.75\textwidth]{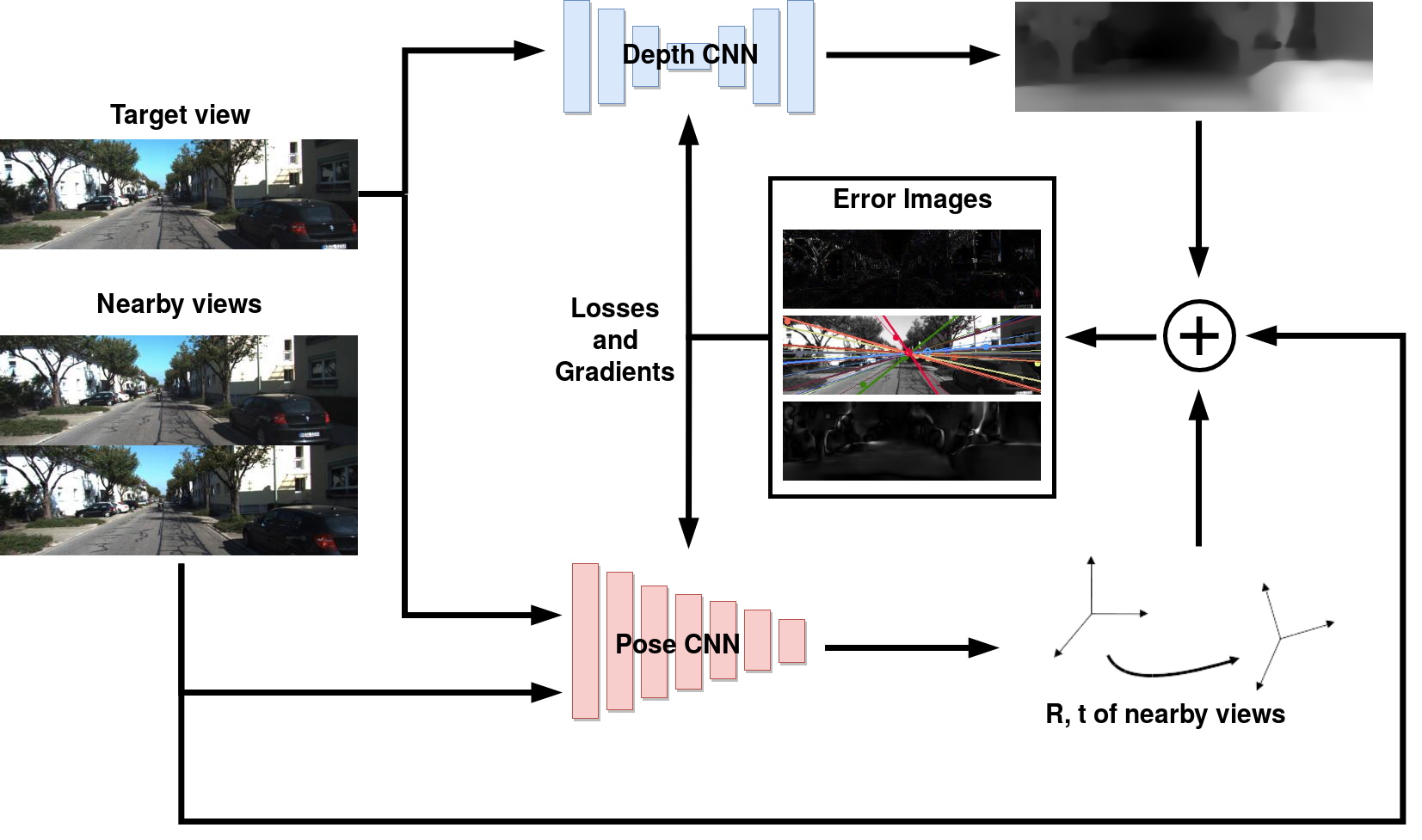}
		\caption{\small{Overview of the training procedure. The Depth CNN predicts the inverse depth for a target view. The Pose CNN predicts the relative poses of the source views from the target. The source views are warped into the target frame using the relative poses and the scene depth and the photometric errors between multiple source-target pairs are minimized. These are weighted by the per-pixel epipolar loss calculated using the Essential Matrix obtained from the Five-Point Algorithm \cite{nister2004efficient}.}}
		\label{fig:flow}
		\vspace{-2em}
	\end{center}
\end{figure*}

In our approach, we use 2 CNNs, one for inverse depth prediction, which takes a single image as input, and one for relative pose prediction which takes an image sequence as input. The first image is the target view with respect to which the poses of the other images are calculated. Both the networks are independent of each other but are trained jointly so that the coupling between scene depth and camera motion can be effectively captured. 
The main idea behind the training, similar to that of previous works, is that of ensuring proper scene reconstruction between the source and target views based on the predicted depth and poses. We warp the source view in the target frame and minimize the photometric error between the synthesized image and the target image to ensure that the predicted depth and pose can recreate the scene effectively. Details about this warping process is given in Sec. \ref{ssec:warp} and \ref{ssec:vs}. 

For now, we consider mostly static scenes, i.e. where objects in the scene are rigid. SfMLearner predicts an "explainability mask" along with the pose, which 
denotes the contribution of a pixel to the loss such that pixels of non-rigid objects have a low weight. 
Instead, we use the epipolar loss to weight the pixels as explained in Sec. \ref{ssec:epipolar_losses}. Further explanations about the image warping and losses used are given below. In all the below equations, we use $N$ to denote the total number of pixels in the image.

\subsection{Image Warping}
\label{ssec:warp}
Given a pixel $p$ in normalized coordinates, we calculate its 3D coordinates using its depth $D(p)$. We then transform it into the source frame using the relative pose and project it onto the source image's plane.

\vspace{-0.8em}
\begin{equation}
\label{eq:pixel_warping}
\hat{p} = K(R_{t \rightarrow s}D(p)K^{-1}p + t_{t \rightarrow s})
\vspace{-0.2em}
\end{equation}

where $K$ is the camera calibration matrix, $D(p)$ is the depth of pixel $p$, $R_{t \rightarrow s}$ and $t_{t \rightarrow s}$ are the rotation and translation respectively from the target frame to the source frame. The homogeneous coordinates of $\hat{p}$ are continuous while we require integer values. Thus, we interpolate the values from nearby pixels, using bi-linear sampling, proposed by \cite{jaderberg2015spatial}, as explained in Sec. \ref{ssec:dibr}.\par

\subsection{Novel View Synthesis}
\label{ssec:vs}
We use novel view synthesis using depth image based warping as the main supervisory signal for our networks. Given the per-pixel depth and the relative poses between images, we can synthesize the image of the scene from a novel viewpoint. We minimize the photometric error between this synthesized image and the actual image from the given viewpoint. If the pose and depth are correctly predicted, then the given pixel should be projected to its corresponding pixel in the image from the given viewpoint. \par

Given a target view $I_t$ and $N$ source views $I_s$, we minimize the photometric error between the target view and the source view warped into the target's frame, denoted by $\hat{I_s}$. Mathematically, this can be described by Eq. \ref{eq:warp_loss}

\vspace{-1em}
\begin{equation}
\label{eq:warp_loss}
L_{warp} = \frac{1}{N}\sum_s\sum_p|I_t(p) - \hat{I_s}(p)|
\end{equation}




\subsection{Spatial Smoothing}
\label{ssec:smooth}
In order to tackle the issues of learning wrong depth values for texture-less regions, we try to ensure that the depth prediction is derived from spatially similar areas. One more thing to note is that depth discontinuities usually occur at object boundaries. We minimize $L1$ norm of the spatial gradients of the inverse depth $\partial d$, weighted by the image gradient $\partial I$. This is to account for sudden changes in depth due to crossing of object boundaries and ensure a smooth change in the depth values. This is similar to \cite{godard2017unsupervised}.
\begin{equation}
\label{eq:smooth_loss}
L_{smooth} = \frac{1}{N}\sum_p(|\partial_x d(p)|e^{-|\partial_x I(p)|} + |\partial_y d(p)|e^{-|\partial_y I(p)|})
\end{equation}
\vspace{-1.5em}

\subsection{Epipolar Constraints}
\label{ssec:epipolar_losses}
The problem with simply minimizing the photometric error is that it doesn't take ambiguous pixels into consideration, such as those belonging to non-rigid objects, those which are occluded etc. Thus, we need to weight pixels appropriately based on whether they're properly projected or not. One way of ensuring correct projection is by checking if the corresponding pixel $\hat{p}$ lies on its epipolar line or not, according to Eq. \ref{eq:epipolar}.

We impose epipolar constraints using the Essential Matrix obtained from Nist\'er's Five Point Algorithm \cite{nister2004efficient}. This helps ensure that the warped pixels to lie on their corresponding epipolar line. This epipolar loss $\hat{\Tilde{p}}^TE\Tilde{p}$ is used to weight the above losses, where $E$ is the Essential Matrix obtained using the Five Point Algorithm by computing matches between features extracted using SiftGPU \cite{wu2007siftgpu}.

After weighting, the new photometric loss now becomes 
\vspace{-0.7em}
\begin{equation}
\label{eq:weighted_warp_loss}
L_{warp} = \frac{1}{N}\sum_s\sum_p|I_t(p) - \hat{I_s}(p)|e^{|\hat{\Tilde{p}}^TE\Tilde{p}|}
\end{equation}


The reason behind this is that we wish to ensure proper projection of a pixel, rather than ensure just a low photometric error. If a pixel is projected correctly based on the predicted depth and pose, its epipolar loss would be low. However, for a non-rigid object, even if the pixel is warped correctly, with the pose and depth, the photometric error would be high. A high photometric error can also arise due to incorrect warping. Therefore, to ensure that correctly warped pixels aren't penalized for belonging to moving objects, we weight them with their epipolar distance, thereby giving their photmetric loss a lower weight than those that are wrongly warped.\par
If the epipolar loss is high, it implies that the projection is wrong, giving a high weight to the photometric loss, thereby increasing its overall penalty. This also helps in mitigating the problem of a pixel getting projected to a region of similar intensity by constraining it to lie along the epipolar line.\par

\subsection{Final Loss}
Our final loss function is a weighted combination of the above loss functions summed over multiple image scales.
\begin{equation}
\label{eq:final_loss}
L_{final} = \sum_l (L_{warp}^l  + \lambda_{smooth}L_{smooth}^l)
\end{equation}

where $l$ iterates over the different scale values  and $\lambda_{smooth}$ 
is the coefficient giving the relative weight for the smoothness loss. 
Note that we don't minimize the epipolar loss but use it for weighting the other losses. This way the network tries to implicitly minimize the projection error (the epipolar loss) as it would lead to a reduction in the over all loss. Along with this it also minimizes the result of the projection as well (photometric loss).

\section{Implementation Details}
\subsection{Neural Network Design}
\label{ssec:networks}

The Depth network is inspired from DispNet\cite{mayer2016large}. It has a design of a convolutional-deconvolutional encoder-decoder network with skip connections from previous layers. The input is a single RGB image. We perform prediction at 4 different scales. We normalize the inverse-depth prediction to have unit mean, similar to what is done in the key frames of LSD-SLAM \cite{engel2014lsd} and in \cite{wang2018learning}. \par 

We modify the pose network proposed by \cite{zhou2017unsupervised} by removing their "explainability mask" layers thereby using lesser parameters yet giving better performance. The target view and the source views are concatenated along the colour channel giving rise to an input layer of size $H \times W \times 3N$ where $N$ is the number of input views.  The network predicts 6 DoF poses for each of the $N - 1$ source views relative to the target image. \par 

We use batch normalization\cite{ioffe2015batch} for all the non-output layers. Details of network implementations are given in the appendix in Fig. \ref{fig:networks}.

	\begin{figure*}[h!]
		\centering
		\setlength{\tabcolsep}{1pt}
		\hspace{-0.5em}
		\begin{tabular}{cccc}
			\textbf{Image}                               & \textbf{Ground truth}                      & \textbf{SfMLearner} & \textbf{Proposed Method} \\ 
			\includegraphics[width=0.22\textwidth]{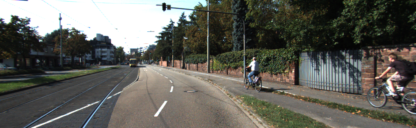} & \includegraphics[width=0.22\textwidth]{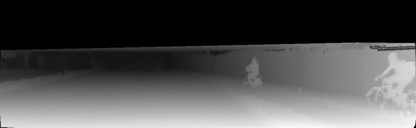} & \includegraphics[width=0.22\textwidth]{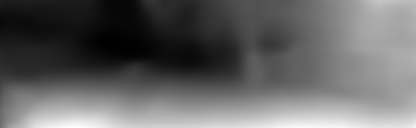} & \includegraphics[width=0.22\textwidth]{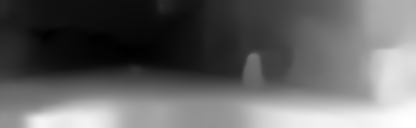} \\
			\includegraphics[width=0.22\textwidth]{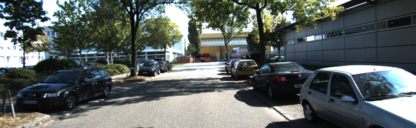} & \includegraphics[width=0.22\textwidth]{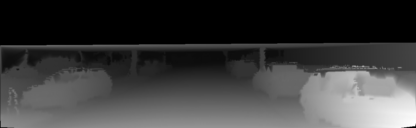} & \includegraphics[width=0.22\textwidth]{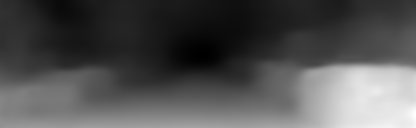} & \includegraphics[width=0.22\textwidth]{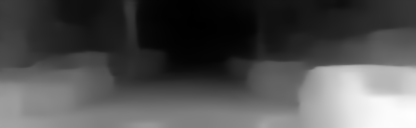} \\
			\includegraphics[width=0.22\textwidth]{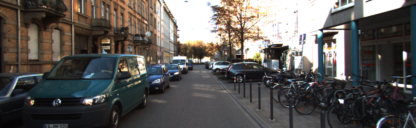} & \includegraphics[width=0.22\textwidth]{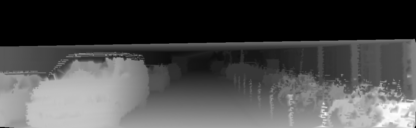} & \includegraphics[width=0.22\textwidth]{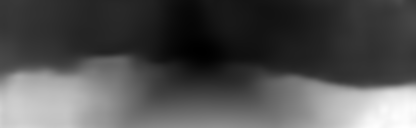} & \includegraphics[width=0.22\textwidth]{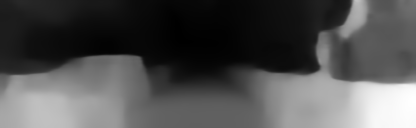} \\
			\includegraphics[width=0.22\textwidth]{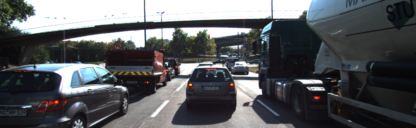} & \includegraphics[width=0.22\textwidth]{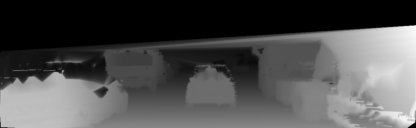} & \includegraphics[width=0.22\textwidth]{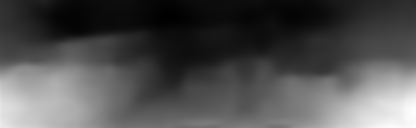} & \includegraphics[width=0.22\textwidth]{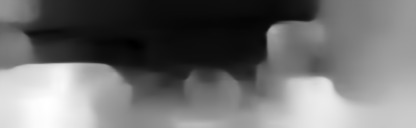} \\
			\includegraphics[width=0.22\textwidth]{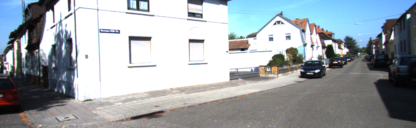} & \includegraphics[width=0.22\textwidth]{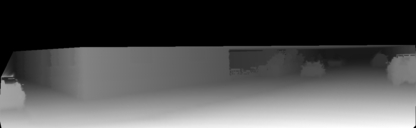} & \includegraphics[width=0.22\textwidth]{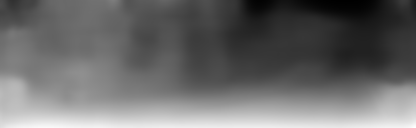} & \includegraphics[width=0.22\textwidth]{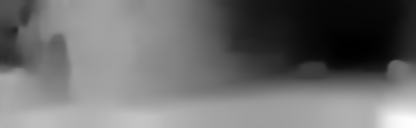} \\
			\includegraphics[width=0.22\textwidth]{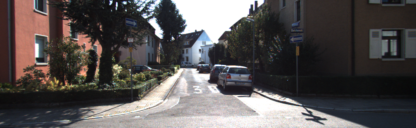} & \includegraphics[width=0.22\textwidth]{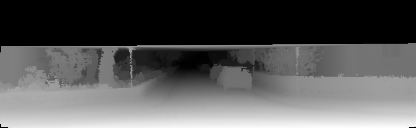} & \includegraphics[width=0.22\textwidth]{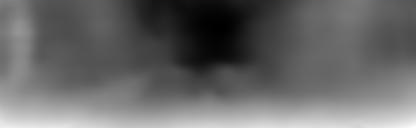} & \includegraphics[width=0.22\textwidth]{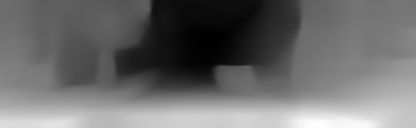} \\
			\includegraphics[width=0.22\textwidth]{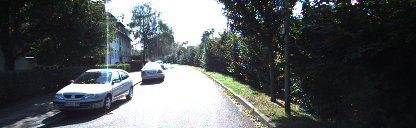} & \includegraphics[width=0.22\textwidth]{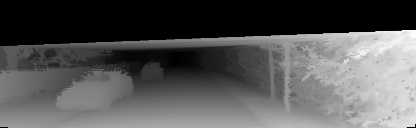} & \includegraphics[width=0.22\textwidth]{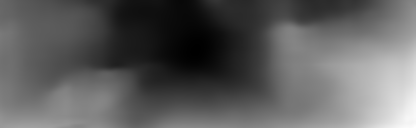} & \includegraphics[width=0.22\textwidth]{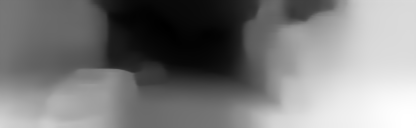} \\
			\arrayrulecolor{red}\hline\\
			\includegraphics[width=0.22\textwidth]{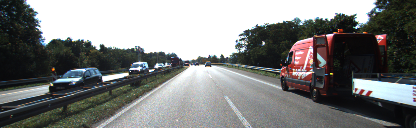} & \includegraphics[width=0.22\textwidth]{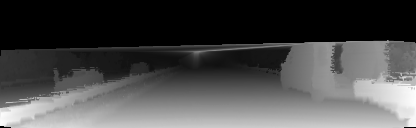} &\includegraphics[width=0.22\textwidth]{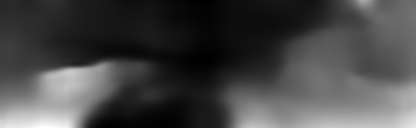} & \includegraphics[width=0.22\textwidth]{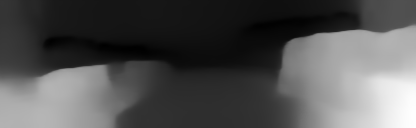} \\
			\includegraphics[width=0.22\textwidth]{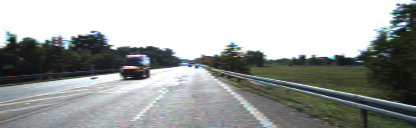} & \includegraphics[width=0.22\textwidth]{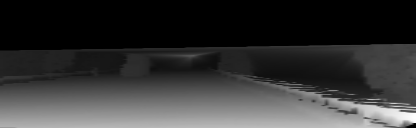} & \includegraphics[width=0.22\textwidth]{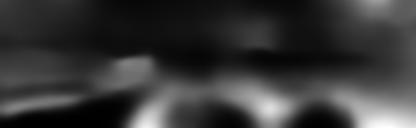} & \includegraphics[width=0.22\textwidth]{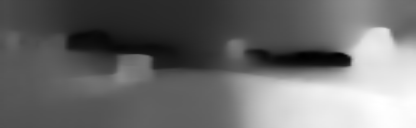} \\
			\includegraphics[width=0.22\textwidth]{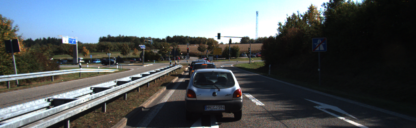} & \includegraphics[width=0.22\textwidth]{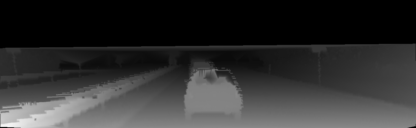} & \includegraphics[width=0.22\textwidth]{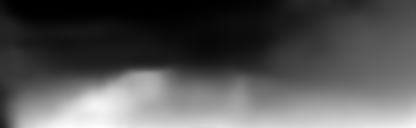} & \includegraphics[width=0.22\textwidth]{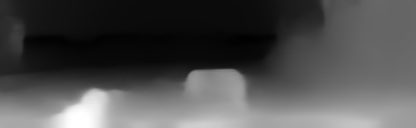}  
			
		\end{tabular}
		\caption{Results of depth estimation compared with SfMLearner. The ground truth is interpolated from sparse measurements. Some of their main failure cases are highlighted in the last 3 rows, such as large open spaces, texture-less regions, and when objects are present right in front of the camera. As it can be seen in the last 3 rows, our method performs significantly better, providing more meaningful depth estimates even in such scenarios. (Pictures best viewed in color.)}
		\label{fig:depth_images}
				\vspace{-1.5em}
	\end{figure*}

\subsection{Training}
\label{ssec:training}
We implement the system using Tensorflow \cite{abadi2016tensorflow}. We use Adam \cite{kingma2014adam} with $\beta1 = 0.9$, $\beta2 = 0.999$, a learning rate of 0.0002 and a mini-batch of size 4 in our training. We set $\lambda_{smooth} = 0.2/2^l$ where $l$ is the scale, ranging from 0 - 3. For training, we use the KITTI dataset\cite{geiger2013vision} with the split provided by \cite{eigen2014depth}, which has about 40K images in total. We exclude the static scenes (acceleration less than a limit) and the test images from our training set. We use 3 views as the input to the pose network during the training phase with the middle image as our target image and the previous \& succeeding images as the source images.

\section{Results}

\subsection{KITTI Depth Estimation Results}
\label{ssec:depth_results}

\begin{table*}[h]
		\centering
		\resizebox{\textwidth}{!}{
			\begin{tabular}{|c|c|c|c|c|c|c|c|c|}
				\hline
				\multirow{2}{*}{\textbf{Method}}           & \multirow{2}{*}{\textbf{Supervision}} & \multicolumn{4}{c|}{\textbf{Error Metric} (lower is better)}     & \multicolumn{3}{c|}{\textbf{Accuracy Metric} (higher is better)} \\
				
				\cline{3-9}
				&                              & Abs. Rel. & Sq. Rel. & RMSE  & RMSE log &   $\delta < 1.25$         &$\delta < 1.25^2$            &     $\delta < 1.25^3$      \\
				
				\hline
				Train set mean                    & --                        & 0.403   & 5.53    & 8.709 & 0.403    & 0.593      & 0.776      & 0.878     \\
				
				Eigen et al. \cite{eigen2014depth} Coarse       & Depth                        & 0.214   & 1.605   & 6.563 & 0.292    & 0.673      & 0.884      & 0.957     \\
				
				Eigen et al. \cite{eigen2014depth} Fine         & Depth                        & 0.203   & 1.548   & 6.307 & 0.282    & 0.702      & 0.89       & 0.958     \\
				
				Liu et al. \cite{liu2016learning}               & Depth                        & 0.202   & 1.614   & 6.523 & 0.275    & 0.678      & 0.895      & 0.965     \\
				
				Godard et al.\cite{godard2017unsupervised}            & Stereo                        & 0.148   & 1.344   & 5.927 & 0.247    & 0.803      & 0.922      & 0.964     \\
				
				
				
				
				\hdashline
				
				DDVO\cite{wang2018learning}            & Mono                        & 0.151   & 1.257   & 5.583 & 0.228    & 0.810      & 0.936      & 0.974     \\	
							
				Yang \textit{et al.}\cite{yang2017unsupervised}           & Mono                        & 0.182   & 1.481   & 6.501 & 0.267    & 0.725      & 0.906      & 0.963     \\
				
				Geonet\cite{yin2018geonet} (ResNet)           & Mono                        & 0.155   & 1.296   & 5.857 & 0.233    & 0.793      & 0.931      & 0.973     \\

				Geonet\cite{yin2018geonet} (updated from github)           & Mono                        & 0.149   & 1.060   & 5.567 & 0.226    & 0.796      & 0.935      & 0.975     \\

				Mahjourian \textit{et al.}\cite{mahjourian2018unsupervised}           & Mono                        & 0.163   & 1.240   & 6.220 & 0.250    & 0.762      & 0.916      & 0.968     \\

				LEGO \cite{yang2018lego} & Mono & 0.162   & 1.352   & 6.276 & 0.252 & 0.783 & 0.921 & 0.969 \\

				SfMLearner (w/o explainability)   & Mono                             & 0.221   & 2.226   & 7.527 & 0.294    & 0.676      & 0.885      & 0.954     \\

				SfMLearner                        & Mono                              & 0.208   & 1.768   & 6.856 & 0.283    & 0.678      & 0.885      & 0.957     \\

				SfMLearner (updated from github)                        & Mono                              & 0.183   & 1.595   & 6.709 & 0.270    & 0.734      & 0.902      & 0.959     \\

				Ours                        & Mono                              & 0.175&	1.396&	5.986&	0.255&	0.756&	0.917&	0.967     \\
				\hline		
				\hline		
				Garg \textit{et al.} \cite{garg2016unsupervised}      &       Stereo                       & 0.169   & 1.08    & 5.104 & 0.273    & 0.740       & 0.904      & 0.962     \\


				\hdashline
				GeoNet\cite{yin2018geonet} (ResNet)           & Mono                        & 0.147   & 0.936   & 4.348 & 0.218    & 0.810      & 0.941      & 0.977     \\

				Mahjourian \textit{et al.}\cite{mahjourian2018unsupervised}           & Mono                        & 0.155   & 0.927   & 4.549 & 0.231    & 0.781      & 0.931      & 0.975     \\

				SfMLearner (w/o explainability) &    Mono                          & 0.208   & 1.551   & 5.452 & 0.273    & 0.695      & 0.900        & 0.964     \\

				SfMLearner                      &         Mono                     & 0.201   & 1.391   & 5.181 & 0.264    & 0.696      & 0.900        & 0.966    \\

				Ours                      & Mono                              & 0.168&	1.105&	4.624&	0.241&	0.773&	0.927&	0.972     \\

				\hline
			\end{tabular}
		}
		\vspace{0.01em}
		\caption{Single View Depth results using the split of \cite{eigen2014depth}. Garg \textit{et al.}\cite{garg2016unsupervised} cap their depth at 50m which we show in the bottom part of the table. The dashed line separates methods that use some form of supervision from purely monocular methods. $\delta$ is the ratio between the scaled predicted depth and the ground truth. Further details about the error and accuracy metrics can be found in \cite{eigen2014depth}. Baseline numbers taken from \cite{wang2018learning,yang2017unsupervised,yin2018geonet,mahjourian2018unsupervised,yang2018every}.}
		\label{table:depth_results}
	\end{table*}

We evaluate our performance on the 697 images used by \cite{eigen2014depth}. We show our results in Table \ref{table:depth_results}. Our method's performance exceeds that of SfMLearner and Yang \textit{et al.}\cite{yang2017unsupervised}, both of which are monocular methods. We also beat methods which use depth supervision \cite{eigen2014depth,liu2016learning}  and calibrated stereo supervision\cite{garg2016unsupervised}. However, we fall short of Godard \textit{et al.}\cite{godard2017unsupervised}, who also use stereo supervision and incorporate left-right consistency, making their approach more robust. \par 

In monocular methods, we fall short of GeoNet\cite{yin2018geonet} and DDVO\cite{wang2018learning}. GeoNet\cite{yin2018geonet} uses a separate encoder decoder network to predict optical flow from rigid flow, which largely increases the no. of parameters. DDVO\cite{wang2018learning}, uses expensive non-linear optimizations on top of the network outputs in each iteration, which is computationally intensive.\par 

Our depth estimates deviate just marginally compared to LEGO\cite{yang2018lego}and Mahjourian \textit{et al.}\cite{mahjourian2018unsupervised}. However, the complexity of our method is much lesser than LEGO\cite{yang2018lego}, which incorporates many complex losses (depth-normal consistency, surface normal consistency, surface depth consistency) and uses a much larger no. of parameters as they predict edges as well. Mahjourian \textit{et al.}\cite{mahjourian2018unsupervised} minimize the Iterative Point Cloud (ICP) \cite{besl1992method,chen1991object,rusinkiewicz2001efficient} alignment error between predicted pointclouds, which is an expensive operation, while ours uses just simple, standard epipolar constraints. This shows that our method being simple, effective and motivated by geometric association between multiple views produces comparable results with methods using complex costs and more no. of parameters. \par

As shown in Fig. \ref{fig:depth_images}, our method performs better where SfMLearner fails, such as texture-less scenes and open regions. This shows the effectiveness of using epipolar geometry, rather training an extra parameter per-pixel to handle occlusions and non-rigidity. We also provide sharper outputs which is a direct result of using an edge-aware smoothness that helps capture the shape of objects in a better manner. We scale our depth predictions to match the ground truth depth's median. Further details about the depth evaluation metrics can be found in \cite{eigen2014depth}.

	\begin{table*}[]
		\centering
		\resizebox{\textwidth}{!}{\begin{tabular}{|c||c|c|c|c|c||c|c|}
\hline
\multirow{2}{*}{\textbf{Seq}} & \multicolumn{5}{c||}{\textbf{Avg. Trajectory Error}} & \multicolumn{2}{c|}{\textbf{Avg. Translational Direction Error}} \\
\cline{2-8}
& \textbf{SfMLearner} & \textbf{GeoNet\cite{yin2018geonet}} & \textbf{LEGO\cite{yang2018lego}}& \textbf{Mahjourian \textit{et al.}\cite{mahjourian2018unsupervised}} & \textbf{Ours} & \textbf{Five Pt. Alg.} & \textbf{Ours} \\
\hline
\textbf{00} & 0.5099$\pm$0.2471 & 0.4975$\pm$0.1791& 0.4973$\pm$0.1893& 0.4955$\pm$0.1727& 0.4969$\pm$0.1775 & 0.0084$\pm$0.0821 & \textbf{0.0040$\pm$0.0156} \\
\textbf{01} & 1.2290$\pm$0.2518 & 1.1474$\pm$0.2177& 1.1665$\pm$0.2261& 1.1365$\pm$0.2136& 1.1433$\pm$0.2156 & 0.0061$\pm$0.0807 & \textbf{0.0032$\pm$0.0074} \\
\textbf{02} & 0.6330$\pm$0.2328 & 0.6492$\pm$0.1796& 0.6476$\pm$0.1858& 0.6552$\pm$0.1809 & 0.6509$\pm$0.1797 & 0.0035$\pm$0.0509 & \textbf{0.0021$\pm$0.0026} \\
\textbf{03} & 0.3767$\pm$0.1527 & 0.3593$\pm$0.1243& 0.3610$\pm$0.1294& 0.3587$\pm$0.1237& 0.3592$\pm$0.1239 & 0.0142$\pm$0.1611 & \textbf{0.0027$\pm$0.0042} \\
\textbf{04} & 0.4869$\pm$0.0537 & 0.6357$\pm$0.0604& 0.6062$\pm$0.0597& 0.6603$\pm$0.0625& 0.6439$\pm$0.0604 & 0.0182$\pm$0.2131 & \textbf{0.0002$\pm$0.0013} \\
\textbf{05} & 0.5013$\pm$0.2564 & 0.4918$\pm$0.1996& 0.4935$\pm$0.2060& 0.4925$\pm$0.1943& 0.4920$\pm$0.1984 & 0.0130$\pm$0.0945 & \textbf{0.0043$\pm$0.0047} \\
\textbf{06} & 0.5027$\pm$0.2605 & 0.5394$\pm$0.1624& 0.5308$\pm$0.1796& 0.5434$\pm$0.1541& 0.5392$\pm$0.1616 & 0.0130$\pm$0.1591 & \textbf{0.0080$\pm$0.0699} \\
\textbf{07} & 0.4337$\pm$0.3254 & 0.4001$\pm$0.2412& 0.4115$\pm$0.2513& 0.3979$\pm$0.2313& 0.3993$\pm$0.2401 & 0.0508$\pm$0.2453 & \textbf{0.0112$\pm$0.0445} \\
\textbf{08} & 0.4824$\pm$0.2396 & 0.4714$\pm$0.1811& 0.4714$\pm$0.1910& 0.4714$\pm$0.1794& 0.4715$\pm$0.1807 & 0.0091$\pm$0.0646 & \textbf{0.0037$\pm$0.0057} \\
\textbf{09} & 0.6652$\pm$0.2863 & 0.6292$\pm$0.2039& 0.6310$\pm$0.2153& 0.6234$\pm$0.1944& 0.6288$\pm$0.2026 & 0.0204$\pm$0.1722 & \textbf{0.0072$\pm$0.0213} \\
\textbf{10} & 0.4672$\pm$0.2398 & 0.4165$\pm$0.1825& 0.4269$\pm$0.1834& 0.4149$\pm$0.1754& 0.4163$\pm$0.1820 & 0.0200$\pm$0.1241 & \textbf{0.0036$\pm$0.0084} \\
\hline
		\end{tabular}
	}
		\caption{Average Trajectory Error (ATE) compared with SfMLearner and Mahjourian \textit{et al.}\cite{mahjourian2018unsupervised} and LEGO \cite{yang2018lego} \& Average Translational Direction Error (ATDE) compared with the Five Point algorithm averaged over 3 frame snippets on the KITTI Visual Odometry Dataset \cite{geiger2012CVPR}. ATE is shown in meters and ATDE, in radians. All values are reported as mean $\pm$ std. dev.}
		\label{table:pose_error}
		\vspace{-1.5em}
	\end{table*}

\subsection{Pose Estimation Results}
\label{ssec:pose_results}

We use sequences 00-10 of the KITTI Visual Odometry Benchmark\cite{geiger2012CVPR}, which have the associated ground truth. We use the same model used for depth evaluation to report our pose estimation results (and of other SOTA methods as well), rather than training a separate model with 5 views (as done by most works). We do so as we believe the same training scheme should suffice to efficiently learn depth and motion, rather than having different sets of experiments for each. We show the Avg. Trajectory Error (ATE) and Avg. Translational Direction Error (ATDE) averaged over 3 frame intervals in Table \ref{table:pose_error}. Before ATE comparison, the scale is corrected to align it with the ground truth. Since ATDE compares only directions, we don't correct the scale. \par 

We perform better on all runs compared to the Five pt. alg. in terms of the ATDE. This is a result of having additional feedback from image warping while the Five pt. alg. uses only sparse point correspondences. \par 
We perform better than SfMLearner\footnote{using the model at \url{github.com/tinghuiz/SfMLearner}} on an avg. showing that epipolar constraints help get better estimates. We perform comparably (difference at scale of $10^{-2}$m) to GeoNet\cite{yin2018geonet}\footnote{using scale normalized model at \url{github.com/yzcjtr/GeoNet}}, Mahjourian \textit{et al}\cite{mahjourian2018unsupervised}\footnote{using model at \url{sites.google.com/view/vid2depth}} and LEGO\cite{yang2018lego}, all of whom use computationally expensive losses and a larger no. of parameters, compared to our method which is much simpler, yet effective and uses lesser no. of parameters. The authors of LEGO\cite{yang2018lego} don't provide a pre-trained model, hence the results shown are after running their model\footnote{\url{github.com/zhenheny/LEGO}} for 24 epochs.

\subsection{Epipolar Loss for discounting motions}

\label{ssec:epipolar_mask}

\begin{figure*}[h!]
		\centering
		\subfigure{\includegraphics[width=0.22\textwidth]{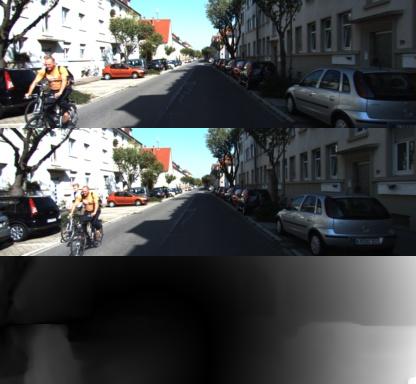}} \subfigure{\includegraphics[width=0.22\textwidth]{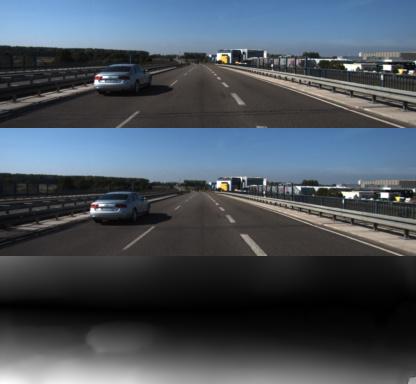}} \subfigure{\includegraphics[width=0.22\textwidth]{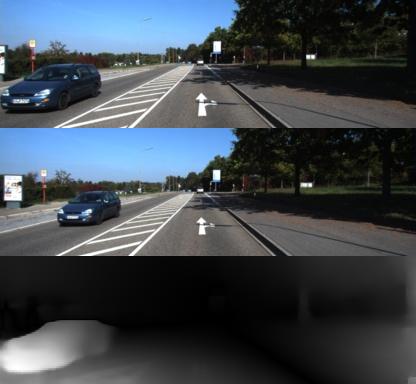}}
		\subfigure{\includegraphics[width=0.22\textwidth]{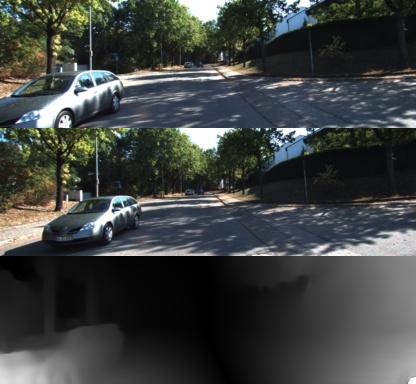}}
		\caption{Epipolar mask visualization. The top is the target image, middle is the source, and bottom is a heatmap of the epipolar weights.}
		\label{fig:epi_mask}
\end{figure*}

We demonstrate the efficacy of using the epipolar loss for discounting ambiguous pixels. As mentioned in the last paragraph of Sec. \ref{ssec:epipolar_losses}, the epipolar loss is useful to give a relatively lower weight to pixels that are properly warped, but belong to non-rigid objects. As seen in Fig. \ref{fig:epi_mask}, the epipolar masks capture the moving objects (cyclists and cars) in the scene effectively, hence they can be used instead of learning motion masks.

\subsection{Cityscapes Depth Estimation}
\vspace{-0.5em}		

\begin{figure*}[h!]
		\centering
		\begin{tabular}{ccc}
			\textbf{Image}                      & \textbf{SfMLearner} & \textbf{Proposed Method} \\ 
			\includegraphics[width=0.22\textwidth]{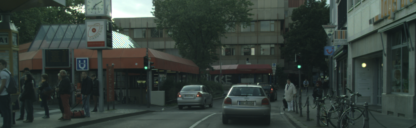} & \includegraphics[width=0.22\textwidth]{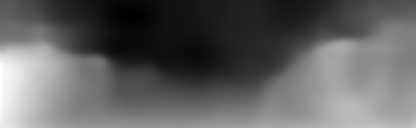} & \includegraphics[width=0.22\textwidth]{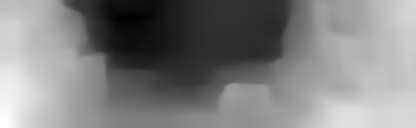} \\
			\includegraphics[width=0.22\textwidth]{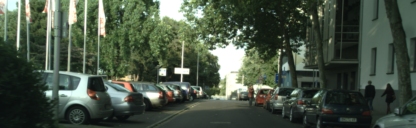} & \includegraphics[width=0.22\textwidth]{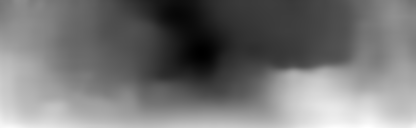} & \includegraphics[width=0.22\textwidth]{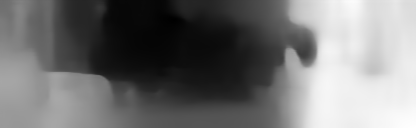} \\
			\includegraphics[width=0.22\textwidth]{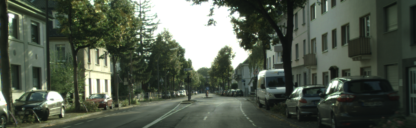} & \includegraphics[width=0.22\textwidth]{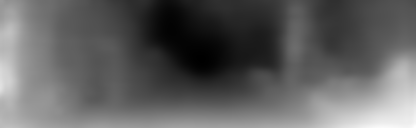} & \includegraphics[width=0.22\textwidth]{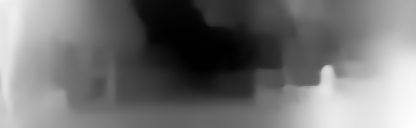} \\
			\includegraphics[width=0.22\textwidth]{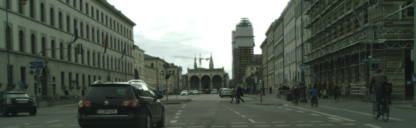} & \includegraphics[width=0.22\textwidth]{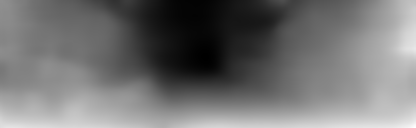} & \includegraphics[width=0.22\textwidth]{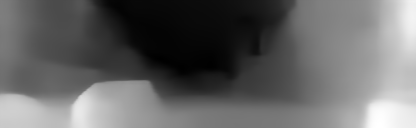} \\
			\includegraphics[width=0.22\textwidth]{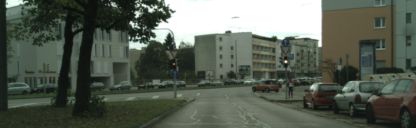} & \includegraphics[width=0.22\textwidth]{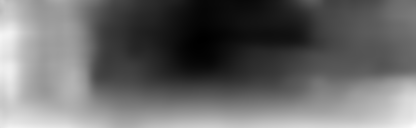} & \includegraphics[width=0.22\textwidth]{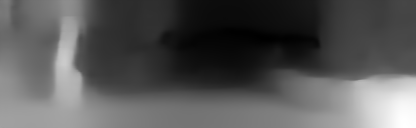} \\
			\includegraphics[width=0.22\textwidth]{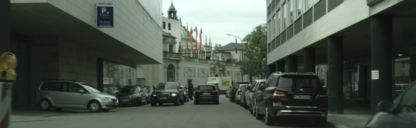} &\includegraphics[width=0.22\textwidth]{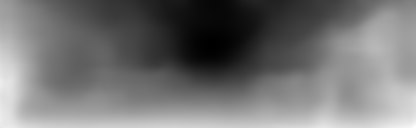} & \includegraphics[width=0.22\textwidth]{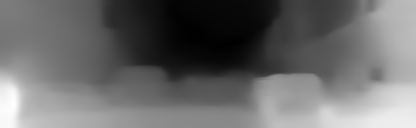} \\
			\includegraphics[width=0.22\textwidth]{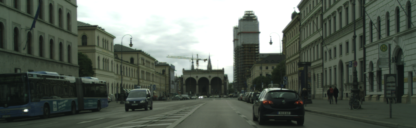} & \includegraphics[width=0.22\textwidth]{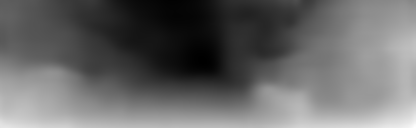} & \includegraphics[width=0.22\textwidth]{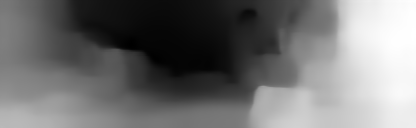} \\
			\includegraphics[width=0.22\textwidth]{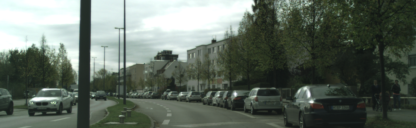} & \includegraphics[width=0.22\textwidth]{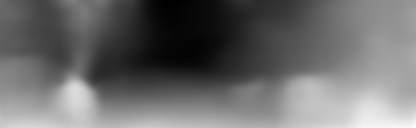} & \includegraphics[width=0.22\textwidth]{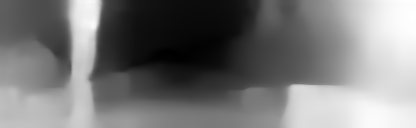} \\
			\includegraphics[width=0.22\textwidth]{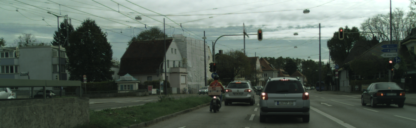} & \includegraphics[width=0.22\textwidth]{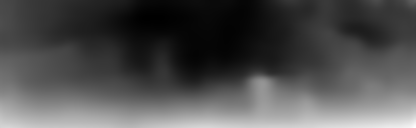} & \includegraphics[width=0.22\textwidth]{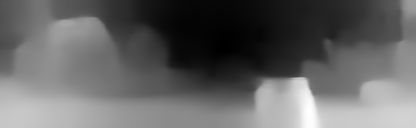} \\
			\includegraphics[width=0.22\textwidth]{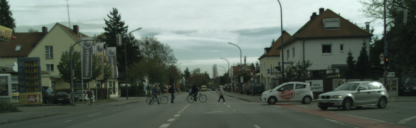} & \includegraphics[width=0.22\textwidth]{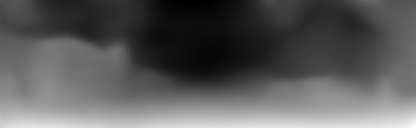} & \includegraphics[width=0.22\textwidth]{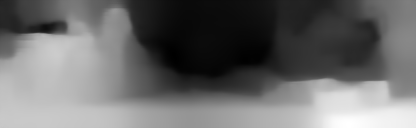} 		
			
		\end{tabular}
		\caption{Results of depth estimation compared with SfMLearner on the Cityscapes dataset. Note that the models that were trained only on the KITTI dataset are tested directly on the Cityscapes dataset without any fine-tuning. These images are previously unseen by the networks. (Pictures best viewed in colour.)}
		\label{fig:depth_images_city}
	\end{figure*}

\label{sec:cityscapes}
We compare our model with SfMLearner (both trained on KITTI) on the Cityscapes dataset \cite{Cordts2015Cvprw,Cordts2016Cityscapes}, which is a similar urban outdoor driving dataset. These images are previously unseen to the networks. As it can be seen in Fig. \ref{fig:depth_images_city}, our model shows better depth outputs on unseen data, showing better generalization capabilities. 

\begin{figure*}[h!]
	\centering
	\begin{tabular}{cccc}
		Image&Ground Truth&SfMLearner&Ours\\
		\includegraphics[width=0.22\textwidth]{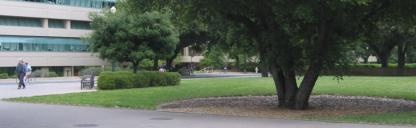} & \includegraphics[width=0.22\textwidth]{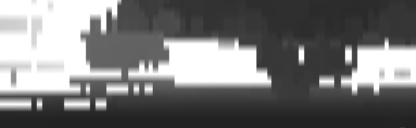} & \includegraphics[width=0.22\textwidth]{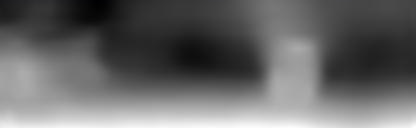} &
		\includegraphics[width=0.22\textwidth]{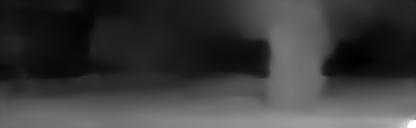} \\
		
		\includegraphics[width=0.22\textwidth]{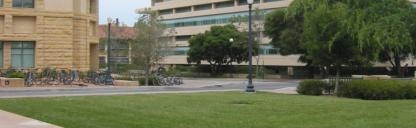} & \includegraphics[width=0.22\textwidth]{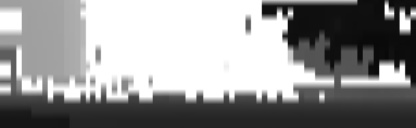} & \includegraphics[width=0.22\textwidth]{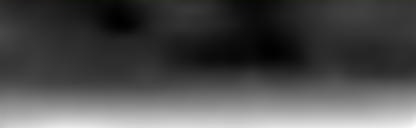} &
		\includegraphics[width=0.22\textwidth]{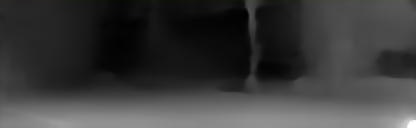} \\
		
		\includegraphics[width=0.22\textwidth]{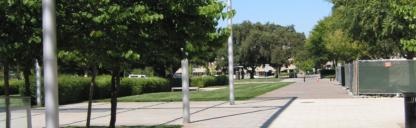} & \includegraphics[width=0.22\textwidth]{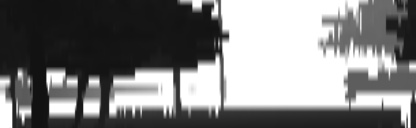} & \includegraphics[width=0.22\textwidth]{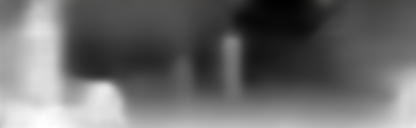} &
		\includegraphics[width=0.22\textwidth]{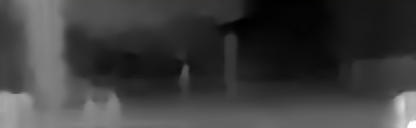} \\
		
		\includegraphics[width=0.22\textwidth]{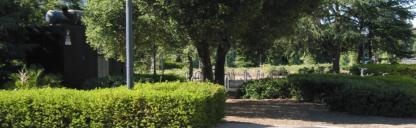} & \includegraphics[width=0.22\textwidth]{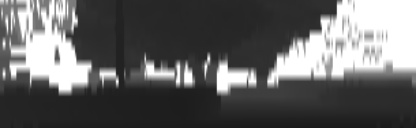} & \includegraphics[width=0.22\textwidth]{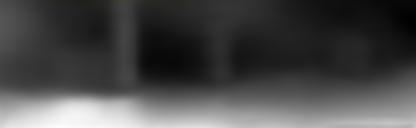} &
		\includegraphics[width=0.22\textwidth]{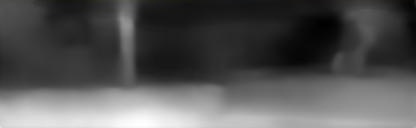} \\
		
		\includegraphics[width=0.22\textwidth]{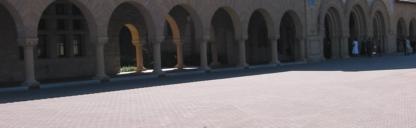} & \includegraphics[width=0.22\textwidth]{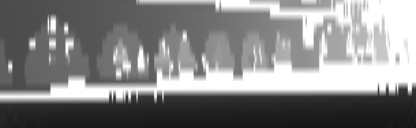} & \includegraphics[width=0.22\textwidth]{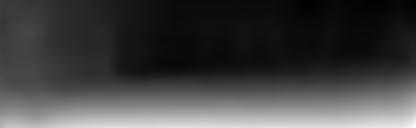} &
		\includegraphics[width=0.22\textwidth]{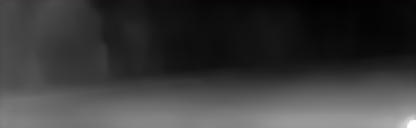} \\
		
		\end{tabular}
		\caption{Results of our model trained on KITTI tested with Make3D}
		\label{fig:make3d}
		\end{figure*}

\begin{table*}[h!]
	\centering
	\resizebox{\textwidth}{!}{
		\begin{tabular}{|c|c|c|c|c|c|c|c|}
			\hline
			\multirow{2}{*}{\textbf{Method}}           & \multicolumn{4}{c|}{\textbf{Error Metric} (lower is better)}     & \multicolumn{3}{c|}{\textbf{Accuracy Metric} (higher is better)} \\
			\cline{2-8}
			& Abs. Rel. & Sq. Rel. & RMSE  & RMSE log &   $\delta < 1.25$         &$\delta < 1.25^2$            &     $\delta < 1.25^3$      \\

			\hline

			SfMLearner (w/o explainability)                                & 0.221   & 2.226   & 7.527 & 0.294    & 0.676      & 0.885      & 0.954     \\
			Ours (only epi)                                                      & 0.217 & 1.639	& 6.746	& 0.297	& 0.651	& 0.875	& 0.954	     \\
			Ours  (only depth-norm)                                                      & 0.187 & 1.726	& 6.530& 0.272	& 0.727	& 0.904	& 0.960	     \\
			Ours (final)                                                      & \textbf{0.175}&	\textbf{1.396}&	\textbf{5.986}&	\textbf{0.255}&	\textbf{0.756}&	\textbf{0.917}&	\textbf{0.967}     \\ 
			\hline
		\end{tabular}
	}
	
	\caption{Ablative study on the effect of different losses in depth estimation. We show our results using the split of \cite{eigen2014depth} while removing the proposed losses. We compare our method with  SfMLearner (w/o explainability) which is essentially similar to stripping our method of the proposed losses and using a different smoothness loss.}
	\label{table:ablation_results}
\end{table*}\par

\subsection{Make3D Depth Estimation}

We compare our model trained on KITTI with SfMLearner, on Make3D \cite{saxena2009make3d}, a collection of still outdoor non-street images, unlike KITTI or Cityscapes. We center-crop the images before predicting depth. Fig. \ref{fig:make3d} shows that our method produces better depth outputs on previously unseen images that are quite different from training images showing that it effectively generalizes to unseen data.

\section{Ablation Study}
We perform an ablation study on the depth estimation by considering variants of our proposed approach. The baseline for comparison is SfMLearner w/o explainability, which is equivalent to removing our additions and having a simpler 2nd order smoothness loss.
We first see the effect of adding just the epipolar constraints, denoted by "Ours (only epi)". We then see how adding just the inverse-depth normalization "Ours (only depth-norm)" affects the performance.  Finally we combine all the proposed additions giving rise to our final loss funtion, denoted by "Ours (final)". 

The results of the study can be seen in Table \ref{table:ablation_results}. All our variants perform significantly better than SfMLearner showing that our method has a positive effect on the learning. Just adding the epipolar constraints improves the results over SfMLearner. The inverse-depth normalization gives a significant improvement as it constrains the depth to lie in a suitable range, which would otherwise cause the inverse-depth to decrease over iterations, leading to a decrease in the smoothness loss, as explained in \cite{wang2018learning}. Finally, combining them all, produces better results than either of the additions individually, showing the effectiveness of our proposed method.\par

\section{Conclusion and Future Work}
\label{sec:conc}
We improve upon a previous unsupervised method for learning monocular visual odometry and single view depth estimation while using lesser number of trainable parameters in a simplistic manner. Our method is able to predict sharper and more accurate depths as well as better pose estimates. Other than just KITTI, we also show better performance on Cityscapes and Make3D as well, using the model trained with just KITTI. This shows that our method generalizes well to unseen data. With just a simple addition, we are able to get performance similar to state-of-the-art methods that use complex losses.

The current method however only performs pixel level inferences. A higher scene level understanding can be obtained by integrating semantics to get better correlation between objects in the scene and depth \& ego-motion estimates, similar to semantic motion segmentation \cite{haque2017joint,haque2017temporal}. \par

Architectural changes could be leveraged by performing multi-view depth prediction to learn how the depth varies over multiple frames or having a single network for both pose and depth to help capture the complex relation between camera motion and scene depth. 


{\small
	\bibliographystyle{ieee}
	\bibliography{references}
}
\clearpage
\appendix
\setcounter{figure}{0}
\renewcommand\thefigure{A\arabic{figure}}
\begin{appendix}
	\section{Neural Network Architecture}	
	\label{sec:network_arch}
	
	\input{nets.tex}

\end{appendix}

\end{document}